\documentclass{INTERSPEECH2023}

\usepackage{xcolor}
\usepackage{algorithm}
\usepackage{algpseudocode}
\usepackage{colortbl,xcolor}
\usepackage{fdsymbol}
\interspeechcameraready

\title{MMER: Multimodal Multi-task Learning for Speech Emotion Recognition}

\name{Sreyan Ghosh$^{\spadesuit}$,\qquad Utkarsh Tyagi$^{\spadesuit}$,\qquad S Ramaneswaran$^{\clubsuit}$,\qquad Harshvardhan Srivastava$^{\varheartsuit}$,\qquad Dinesh Manocha$^{\spadesuit}$}
\address{
  $^{\spadesuit}$University of Maryland, College Park, USA,\\
  $^{\clubsuit}$NVIDIA, Bangalore, India,
  $^{\varheartsuit}$IIT Delhi, India}
\email{\{sreyang,utkarsht,dmanocha\}@umd.edu}

\begin{document}

\maketitle
 
\begin{abstract}
In this paper, we propose \textbf{MMER}, a novel \underline{M}ultimodal  \underline{M}ulti-task learning approach for Speech \underline{E}motion \underline{R}ecognition. MMER leverages a novel multimodal network based on early-fusion and cross-modal self-attention between text and acoustic modalities and solves three novel auxiliary tasks for learning emotion recognition from spoken utterances. In practice, MMER outperforms all our baselines and achieves state-of-the-art performance on the IEMOCAP benchmark. Additionally, we conduct extensive ablation studies and results analysis to prove the effectiveness of our proposed approach. \footnote{Code: https://github.com/Sreyan88/MMER.}
\end{abstract}
\noindent\textbf{Index Terms}: speech emotion recognition, human-computer interaction



\section{Introduction}

In addition to the explicit messages humans convey, they implicitly express emotions in conversations. Speech Emotion Recognition (SER) aims to identify these implicit emotions from spoken human utterances, which proves to be one of the key components of better human-computer interaction systems. 


SER is a well-studied problem in the literature, with a variety of systems proposed that achieve state-of-the-art (SOTA) performance on benchmark datasets \cite{cai2021speech,morais_self,padi2022multimodal,baevski2020wav2vec}. However, most of these systems are uni-modal and learn only from the acoustic modality \cite{wuetal,sajjadetal,luetal1}, with very few systems taking a multi-modal approach \cite{padi2022multimodal,9688036}. We emphasize the importance of multi-modal learning for SER due to the real-world multi-modal nature of emotional expression in humans, which includes body language, facial expressions, word choice, tone of voice, and more. With Automatic Speech Recognition (ASR) systems achieving near-optimal results, we hypothesize that the modality of text is available as a complementary signal to speech and can significantly improve SER performance by eliminating the natural prosodic bias in spoken utterances.

While work on SER has proposed various learning paradigms that achieve great performance \cite{cai2021speech}, we hypothesize that SER can primarily benefit the most by learning to solve auxiliary tasks that can help infuse extra knowledge into the model. For example, the current SOTA on the IEMOCAP benchmark \cite{busso2008iemocap} solves an ASR task with emotion classification, which helps the model learn strong linguistic cues. However, multitask learning is an under-explored area in SER literature, and we emphasize that better auxiliary learning tasks can help the model learn improved representations, thereby improving final SER performance.
\vspace{1mm}

{\noindent \textbf{Main Contributions.}} In this paper, we propose MMER, a novel multimodal multitask learning approach for SER. MMER first leverages a \textit{novel multimodal neural network architecture} to capture fine-grained multimodal emotional information from acoustic and text modalities using speech and its corresponding text transcripts. Our proposed architecture captures fine-grained inter-modality interactions and alleviates unimodal biases. Specifically, the model uses strong contextual representations from self-supervised (SSL) models and learns implicit temporal alignments between both modalities using a novel multimodal interaction module, which we discuss in Section \ref{subsec:multimodal}. Next, MMER solves \textit{three auxiliary tasks} in addition to emotion classification. First, it solves an ASR task by minimizing the CTC loss \cite{graves2006connectionist} to learn the natural monotonic alignment between speech and text and the semantic and syntactic information hidden in the text. Next, we propose to solve two additional contrastive learning-based tasks: (1) \textit{Supervised Contrastive Learning} (SCL): To enforce the model to learn better emotion features from multi-modal data, we solve a supervised contrastive learning task \cite{khosla2020supervised}, where we learn instance discrimination with model representations based on ground-truth instance labels. Specifically, for SCL, instances with the same emotion label make up the positives, and those with different labels, make up the negatives. (2) \textit{Augmented Contrastive Learning} (AGL): To make the model more robust to the data and enforce learned features to be speaker invariant, we augment the text using back-translation and generate speech from that text using a speaker-conditioned TTS, conditioned on a different speaker from the training corpus. In practice, MMER achieves SOTA results on the IEMOCAP benchmark for SER. We also perform extensive analysis and ablation studies to prove the effectiveness of each individual component in MMER.

\begin{figure*}[t]
\centering
\includegraphics[width=2.1\columnwidth]{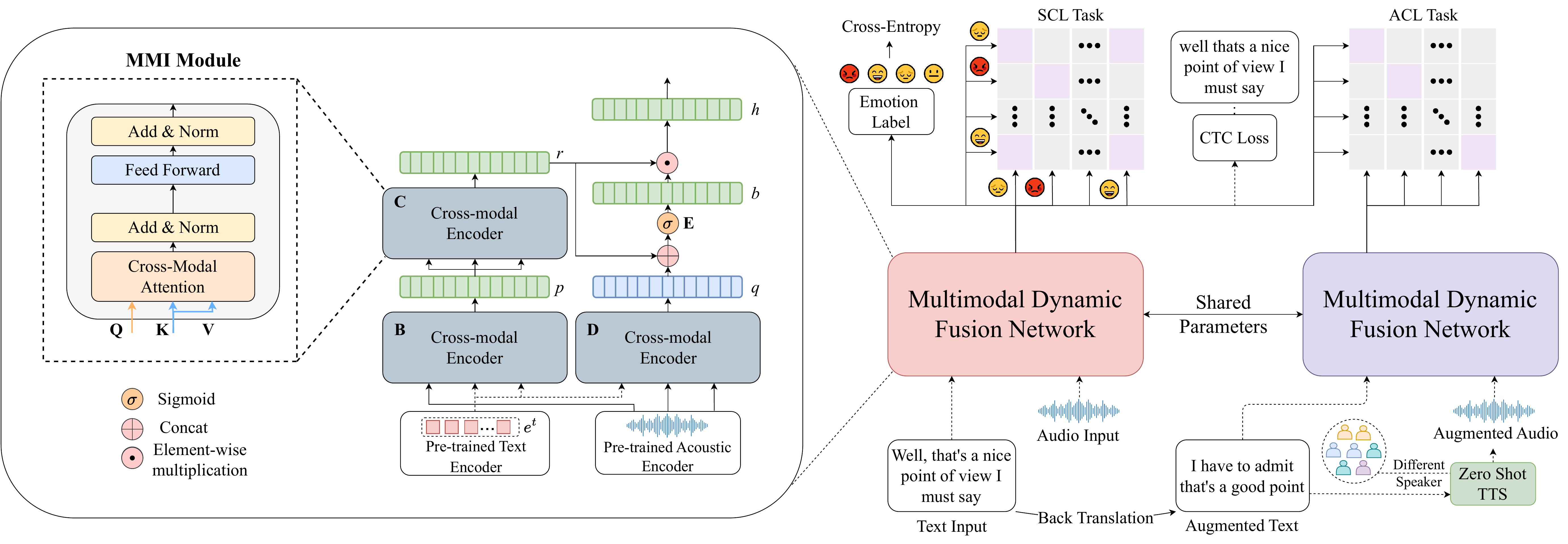}
\caption{\small Illustration of our proposed MMER. MMER introduces a novel Multimodal Dynamic Fusion Network and jointly optimizes 4 different tasks to learn various aspects of SER. We highlight the Multimodal Interaction Module on the left.} 
\label{fig:mmer}
\end{figure*}

\section{Related Work}
{\noindent \textbf{Unimodal SER.}} Uni-modal speech-only SER is the most commonly studied system in literature for SER. Early research focused on extracting low-level features like Mel-frequency cepstral coefficients (MFCCs) and Filter Banks (FBanks) or hand-engineered features like speaker rate, voice quality, etc. These features were then fed to machine learning classifiers, which proved to perform relatively well in terms of classification accuracy. Thanks to deep learning, deep neural networks have achieved a considerable boost in SER performance and can handle raw waveforms or low-level features directly without the need for hand-engineered features \cite{sarma2018emotion,keesing2021acoustic}. With recent advances in self-supervised learning (SSL), pre-trained SSL features, like Natural Language Processing (NLP) \cite{devlin2018bert} have achieved state-of-the-art (SOTA) performance in various downstream speech processing tasks like Automatic Speech Recognition (ASR), Phone Recognition (PER), Speaker Identification (SID), etc. A comprehensive study can be found here \cite{yang21c_interspeech}. The current state-of-the-art on SER \cite{cai21b_interspeech} also uses wav2vec-2.0 as the speech encoder and solves the SER task with ASR as an auxiliary task by minimizing the CTC loss of the network. A recent study also reveals how supervised MTL on SSL pre-trained features can help the performance of a downstream task when the auxiliary task is chosen properly \cite{parthasarathy2018ladder}. 

\vspace{1mm}

{\noindent \textbf{Multimodal SER.}} For multimodal approaches, the most common combination of modalities includes speech and text. Early studies in this area focused on late fusion of multimodal representations \cite{tripathi2018multi,poria2017context,wu2021emotion}. Though this technique is simple and effective at modeling modality-specific interactions, it is not effective at modeling cross-modal interactions \cite{wang2019words}. Early fusion to capture inter-modality interactions has also been explored \cite{sebastian2019fusion}. However, in general, early fusion also suppresses modality-specific interactions and does not outperform late fusion methods in emotion recognition \cite{wang2019words,poria2018multimodal}. To better model the interactions between modalities, researchers have proposed cross-modal attention (CMA) mechanisms \cite{choi2018convolutional,xu2019learning,krishna2020multimodal,chen2021key}. With CMA, features from one modality are allowed to attend to the other, and the interaction between the sequences from the two modalities enables the system to extract the most useful features for emotion recognition. While \cite{choi2018convolutional,xu2019learning} use dot-product attention, very recently, the use of self-attention-based CMA mechanisms for ER has been gaining traction \cite{krishna2020multimodal,pan2020multi,chen2021key}.

\section{Proposed Methodology}

\subsection{Problem Formulation}
Suppose we have a dataset $D$ with $N$ utterances \{$u$\textsubscript{$1$}, $u$\textsubscript{$2$}, $u$\textsubscript{$3$}, $\cdots$, $u$\textsubscript{$N$}\} and their corresponding labels \{$y$\textsubscript{$1$}, $y$\textsubscript{$2$}, $y$\textsubscript{$3$}, $\cdots$, $y$\textsubscript{$N$}\}. Here we assume each utterance $u_i$ has both speech cues $a_i$ and text cues $t_i$ available where $u_i$ $\in$ ($a_i$,$t_i$). $t_i$ can be ASR transcripts or human-annotated transcripts. We formulate the task of ER as assigning an emotion label $y_i$ to each utterance $u_i$, where $y_i$ denotes the probability distribution that the utterance belongs to one of the $j$ unique emotions being studied in the dataset.

\subsection{Multimodal Dynamic Fusion Network}
\label{subsec:multimodal}

\subsubsection{Feature Encoder}

{\noindent \textbf{Contextualized Speech Representations.}} To encode speech to obtain high-level contextualized representations, we use a pre-trained wav2vec-2.0 \cite{baevski2020wav2vec} as our raw waveform encoder. We use the pre-trained checkpoint released by Facebook, pre-trained on 960 hours of Librispeech, and use the wav2vec-2.0-\emph{base} architecture for all our experiments. For each raw audio input $a_i$ of utterance $u_i$ wav2vec-2.0 outputs $e^{a_{i}} \in \mathbb{R}^{J\times768}$ where $J$ depends on the length of the raw audio file and the CNN feature extraction layer of wav2vec-2.0, which extracts frames with a stride of 20ms and a hop size of 25ms.
\vspace{0.5mm}


{\noindent \textbf{Contextualized Token Representations.}} We use RoBERTa\textsubscript{\emph{BASE}} from the transformers family as our contextualized text encoder to encode the transcript of the utterance and obtain rich contextualized token representations. For a total of $M$ tokens, RoBERTa outputs representations $e^{t_{i}}$ $\in$ $\mathbb{R}^{M\times768}$. We use RoBERTa only as a feature extractor and do not train it while fine-tuning our model.



\subsubsection{Multimodal Interaction Module}
Our Multimodal Interaction Module (MMI) consists of 3 Cross Modal Encoder (CME) blocks annotated as $B$, $C$, and $D$ in Fig. \ref{fig:mmer}. Each of these 3 CME blocks is constructed like a generic transformer layer \cite{vaswani2017attention}, where each layer is composed of an \emph{h}-head $\mathbf{CMA}$ module \cite{tsai2019multimodal}, residual connections, and feed-forward layers. In this section, we discuss the working of each of the 3 CME blocks and the acoustic gate $E$ in detail.
\vspace{1mm}

{\noindent \textbf{Speech-Aware Word Representations.}} As shown in Fig.\ref{fig:mmer}, to learn better token representations with the guidance of the associated spoken utterance, we feed wav2vec-2.0 embeddings $\mathbf{A}$ $\in$ $\mathbb{R}^{d \times J}$ as queries and RoBERTa embeddings $\mathbf{T}$ $\in$ $\mathbb{R}^{d \times M}$ as keys and values into the $\mathbf{CMA}$ module of $\mathbf{CME}$ block $B$ as follows:

\begin{equation}
    \mathbf{CMA}(\mathbf{A}, \mathbf{T})=\operatorname{softmax}\left(\frac{\left[\mathbf{W}_{\mathbf{q}_{\mathbf{i}}} \mathbf{A}\right]^{\top}\left[\mathbf{W}_{\mathbf{k}_{\mathbf{i}}} \mathbf{T}\right]}{\sqrt{d / m}}\right)\left[\mathbf{W}_{\mathbf{v}_{\mathbf{i}}} \mathbf{T}\right]^{\top}
\end{equation}

where \{$\mathbf{W}_{\mathbf{q}_{\mathbf{i}}}$, $\mathbf{W}_{\mathbf{k}_{\mathbf{i}}}$, $\mathbf{W}_{\mathbf{v}_{\mathbf{i}}}$\} $\in$ $\mathbb{R}^{d/m \times h}$ denote the query, key, and value weight matrices, respectively, for the $i^{th}$ attention head. The final output representation of the $\mathbf{CME}$ block $B$ is now $\mathbf{P}$ $=$ ($\mathbf{p_0}$,$\mathbf{p_1}$, $\cdots$, $\mathbf{p_{m-1}}$). Next, to address the fact that each generated representation $\mathbf{p}_i$ in the previous block corresponds to the $i^{th}$ acoustic embedding and not the token embedding, we feed $\mathbf{P}$ to another $\mathbf{CME}$ block $C$, which treats the original RoBERTa embeddings $\mathbf{T}$ as queries and $\mathbf{P}$ as keys and values. Finally, we obtain the final Speech-Aware Word Representations as $\mathbb{R}$ $=$ ($\mathbf{r_0}$,$\mathbf{r_1}$, $\cdots$, $\mathbf{r_{j-1}}$).
\vspace{1mm}

{\noindent \textbf{Word-Aware Speech Representations.}} To obtain the word-aware speech representations and align each word to its closely related frame or wav2vec-2.0 embeddings, we make use of another $\mathbf{CME}$ block $D$ by treating $\mathbf{T}$ as queries and $\mathbf{A}$ as keys and values. The final representations obtained from the block can be denoted as $\mathbf{Q}$ $=$ ($\mathbf{q_0}$,$\mathbf{q_1}$, $\cdots$, $\mathbf{q_{j-1}}$). Phoneme alignment has been long studied in speech science and acoustics, and we hypothesize that this step is important so that each word can assign relative importance to the frames or embeddings important or not important to it.
\vspace{1mm}

{\noindent \textbf{Acoustic Gate.}} Speech frames might encode redundant information like random noise and other redundant speech cues. Thus, it is important to implement an acoustic gate $E$ that can dynamically control the contribution of each speech frame embedding. Following previous work, we implement an acoustic gate $\mathbf{g}$ as follows:

\begin{equation}
\mathbf{g}=\sigma\left(\mathbf{W}_{g}^{\top}[ \mathbf{R} ; \mathbf{Q}] + \mathbf{B}_{g}\right)
\end{equation}

where $\mathbf{W}_{g}$ $\in$ $\mathbb{R}^{2dXd}$ is a weight matrix, $\mathbf{B}_{g}$ $\in$ $\mathbb{R}^{d}$ is the bias, and $\sigma$ is the element-wise sigmoid function. Finally, based on the gate output, the final word-aware speech representations are obtained by $\mathbf{Q}$ = $\mathbf{g} . \mathbf{Q}$. After this step, we concatenate the speech-aware word representations and word-aware speech representations to obtain our final cross-modal MMI representations $\mathbf{M}$ $\in$ $\mathbb{R}^{2d}$ where $\mathbf{M}$ = [$\mathbf{Q}$ ; $\mathbf{R}$], and pass it through a linear transformation $l(.)$, which down-projects $\mathbf{M}$ to a $d$ dimensional space.

\subsection{Multi-task Learning}
As discussed earlier, MMER solves a total of 4 tasks for learning SER. In this sub-section, we briefly discuss all these losses and their contribution to learning SER.
\vspace{0.5mm}

{\noindent \textbf{Cross-Entropy Loss.}} Cross-Entropy is the most common loss used for learning SER. To calculate the Cross-Entropy Loss with ground-truth emotion labels, we first employ max pooling $\mathbf{mp(.)}$ over wav2vec-2.0 speech representations ($\mathbf{A}$) and MMI module ($\mathbf{M}$) independently across the time-step axis and then concatenate the embeddings to obtain a single final embedding $\mathbb{R}^{2d}$.  This final embedding is then passed through a linear transformation and softmax activation function as follows:

\begin{equation}
\label{eqn:softmax}
\hat{y}=\mathbf{softmax}\left(\mathbf{W}_{p}^{\top}[ \mathbf{mp}(\mathbf{A}) ; \mathbf{mp}(\mathbf{M})] + \mathbf{B}_{p}\right)
\end{equation}

where $\hat{y}$ $\in$ $\mathbb{R}^{4}$ is the single vector representation for each utterance, $\mathbf{W}_{p}$ $\in$ $\mathbb{R}^{2d \times 4}$ is a weight matrix, $\mathbf{softmax(.)}$ denotes the softmax activation function, and $\mathbf{mp(.)}$ denotes the attention pooling operation across the embedding axis. After this step, Cross Entropy is calculated by $\mathcal{L}_{\mathrm{CE}}=\operatorname{CrossEntropy}(\hat{y_i}, y_i)$.
\vspace{1mm}

\begin{algorithm}[t]
\scriptsize
\caption{Supervised \& Augmented Contrastive Learning}\label{alg:cap}
\begin{algorithmic}
\Require A list of emotion labels for all data in a batch is L; each emotion is divided into four categories; The Multimodal Dynamic Fusion Network is $\operatorname{MDFN}$; the texts are T; the audio files are A; $\operatorname{BT}$ denotes back-translation, and $\operatorname{TTS}$ denotes zero-shot speaker-conditioned text-to-speech; $\operatorname{E_{speaker}}$ are the speaker embeddings for speakers in batch; $C$ denotes the length of $L_c$; $S$ denotes the length of $L$; $\mathcal{L}_\mathrm{SCL}$ denotes Supervised contrastive loss $\mathcal{L}_\mathrm{ACL}$ denotes augmented contrastive loss.
\State Initialize $L_c = [L - 0, ..., L - 4]$ and $L_t =
list()$
\For{$i = 1; i <= C; i++ $}
    \State initialize $\Tilde{L}_{t} = list()$
    \For{$j = 1; j <= T ; j++ $}
        \If{$L_{c}[i][j]$ equals $0$} 
            \State $\Tilde{L}_{t}.append(j)$
        \EndIf
    \EndFor 
    \State $L_{t}.append(\Tilde{L}_{t})$
\EndFor
\State $R = \operatorname{MDFN}(T, A)$
\State $\Tilde{T} = \operatorname{BT}(T)$
\State $R_{au} = \operatorname{MDFN}(\Tilde{\operatorname{T}}, \operatorname{TTS}(\Tilde{T}, \operatorname{E_{speaker}}))$
\State $\Tilde{l}_{pn} = \operatorname{einsum}(nc, ck \rightarrow nk, [R, RT ])$
\State $l_{pn} = \operatorname{LogSoftmax}(\Tilde{l}_{pn}/ \tau))$
\State $L_{cl} = L_{t}[L[1]]$
\For{$q = 2; q <= S, q + +$}
    \State $L_{cl} = concat(L_{cl}, L_{t}[L[q]] + q × T )$
\EndFor 
\State $\mathcal{L}_\mathrm{SCL}= \operatorname{gather}(l_{pn}, index = L_{cl})/T$
\State $l_{pn} = \operatorname{einsum}(nc, ck \rightarrow nk, [R, RT_{au}])$
\State $cl_{label} = \operatorname{arange}(S)$
\State $\mathcal{L}_\mathrm{ACL} = \operatorname{CrossEntropy}(l_{pn}/ \tau, cl_{label})$
\State \Return $\mathcal{L}_\mathrm{SCL}$ $\mathcal{L}_\mathrm{ACL}$
\end{algorithmic}
\end{algorithm}

{\noindent \textbf{CTC Loss.}} MMER next solves the ASR task by minimizing the CTC loss. Learning to solve the ASR task encourages the model to learn linguistic properties like speech and text by leveraging the natural monotonic alignment between the acoustic and textual modalities. To do this, we first pass the raw un-pooled embeddings $\mathbf{A}$ from the wav2vec-2.0 encoder through a linear layer as follows:

\begin{equation}
\hat{t} = \mathbf{softmax}\left(\mathbf{W}_{c}^{\top}\mathbf{A} + \mathbf{B}_{c}\right)
\end{equation}

where $\hat{t}$ $\in$ $\mathbb{R}^{J \times V}$, $J$ is the number of speech frames output by the wav2vec-2.0 CNN feature extractor, and $V$ is the size of our vocabulary or the number of unique characters and symbols in our corpus and an extra blank token. $\mathbf{W}_{c}$ $\in$ $\mathbb{R}^{d \times V}$ and $\mathbf{B}_{c}$ is the added bias. After this step; we calculate the $\operatorname{CTC}$ loss by $\mathcal{L}_{\mathrm{CTC}}=\operatorname{CTC}(\hat{t},t)$, where $t_i$ $\in$ \{$t_0$,$\cdots$,$t_i$,$\cdots$,$t_N$\} is a pre-processed version of the original $t_i$ where we remove all punctuation and convert all characters to uppercase.
\vspace{1mm}

{\noindent \textbf{Supervised Contrastive Learning.}} Supervised Contrastive Learning (SCL) \cite{khosla2020supervised} supplements the Cross-Entropy to learn better emotion features. Precisely, SCL solves the generic instance discrimination contrastive learning task with multimodal representations $\mathbf{M}$, but in the presence of emotion labels. To solve SCL, we first divide the representations in each batch into multiple subsets according to their emotion label. Then, for each subset, representations within that subset act as the positives, while representations in another subset act as the negatives. Fig. \ref{fig:mmer} illustrates the process, and we show specific steps in Algorithm \ref{alg:cap}.
\vspace{1mm}

{\noindent \textbf{Augmented Contrastive Learning.}} Augmented Contrastive Learning (ACL) encourages MMER to learn invariant features in the data. Past work has shown that SER benefits from learning speaker invariance \cite{9054580}. However, in this work, we take a slightly different approach to consider the multimodal nature of MMER and additionally learn semantic invariance in text. Thus, MMER solves another instance discrimination contrastive learning task between multimodal representations $\mathbf{M}$, where the representations are learned from the augmented text and speech cues. To augment the text, we use back-translation, which refers to translating an existing text into a target language and then back into the source language. Yu \textit{et al.} \cite{yu2018qanet} show that back-translation can generate diverse sentences while preserving the semantics of the original sentence. For speech, our primary objective is to generate an augmented utterance of the same emotion but as if uttered by a different speaker. To perform this, we use a SOTA zero-shot speaker-conditioned TTS \cite{casanova2022yourtts} and generate utterances from the back-translated text. The system proposed by Casanova \textit{et al.} \cite{casanova2022yourtts} takes speaker embeddings as input added to the text, and we calculate these embeddings with all utterances from a speaker randomly sampled from the dataset but expressing a similar emotion. Fig. \ref{fig:mmer} illustrates the process, and we show specific steps in Algorithm \ref{alg:cap}. Finally, for optimizing MMER we minimize $\mathcal{L}$ as: $ \mathcal{L} = \mathcal{L} + \alpha \mathcal{L}_\mathrm{CTC} + \beta \mathcal{L}_\mathrm{SCL} + \gamma \mathcal{L}_\mathrm{ACL}$ where $\alpha$, $\beta$ and $\gamma$ are hyper-parameters that we tune.


\section{Experiments and Results}
\label{sec:experiments}
{\noindent \textbf{Dataset.}} Following much of the prior work in SER literature, we train and evaluate all our models on the IEMOCAP dataset \cite{busso2008iemocap}. IEMOCAP contains approximately 12 hours of speech from a total of 10 speakers, all of which come from 5 scripted sessions, acted by professional actors. To keep our dataset settings consistent with the prior work and for a fair comparison, we evaluate our models on utterances assigned to one of the five emotions (\emph{Happy}, \emph{Angry}, \emph{Neutral}, \emph{Sad} and \emph{Excited}) and merge all samples labeled with \emph{Excited} to \emph{Happy}. For evaluation, we follow the five-fold cross-validation approach, where at each fold, we leave one session out as the test set and take the average of the weighted accuracy obtained at each fold.
\vspace{1mm}

{\noindent \textbf{Baselines.}} We build unimodal baselines with just text and speech modalities, where the text baseline uses RoBERTa\textsubscript{BASE} as the contextualized text encoder, followed by a single linear layer and softmax activation for classification. For the unimodal speech baseline, we use the same setup but replace our encoder with pre-trained wav2vec-2.0-base, pre-trained on 960hrs of LibriSpeech \cite{panayotov2015librispeech}. We also build a naive multimodal baseline where we simply concatenate pooled self-supervised representations $\mathbf{mp}(\mathbf{A})$ and $\mathbf{mp}(\mathbf{T})$ in a single-task SER learning setup. We compare our model with other methods in the literature evaluated on 5-fold cross-validation setups, including unimodal and multimodal approaches. All results for prior art have been taken from the literature (weighted accuracy unless stated otherwise). We only re-implement the current state-of-the-art approach \cite{cai2021speech} under the 5-fold cross-validation setup for a fair comparison.
\vspace{1mm}

{\noindent \textbf{Hyper-parameters.}} Since we use the \emph{base} architectures for both RoBERTa and wav2vec-2.0, our $d$ effectively takes a value of 768. We trained and evaluated all our models with a batch size of 4 and accum-grad of 4 for 100 epochs. For training, we kept the learning rate constant at $1e^{-5}$, which worked well for all our setups. For our multi-task learning setup, we trained our models with $\alpha,\beta,\gamma$ = 0.1 where the search was performed among $\alpha,\beta,\gamma \in \{1,0.1,0.01,0.001\}$ with grid search. Each training and inference step took 10 minutes on a single NVIDIA A100 GPU. MMER has $\approx$228M parameters.
\vspace{1mm}

{\noindent \textbf{Quantitative Analysis.}} Table \ref{tab:results} compares the performance of MMER with all our baselines. MMER achieves SOTA performance on the IEMOCAP benchmark, with the closest being \cite{morais_self}, where the author uses 2 contextualized speech encoders, resulting in more than double the number of parameters as ours. MMER also benefits from minimal trainable parameter addition over \cite{cai2021speech} or a simple wav2vec-2.0. We achieved 75.0\% WA when Google transcripts were used instead of gold transcripts for inference.
\vspace{1mm}

\begin{figure}[t]
\centering
\includegraphics[width=0.45\columnwidth]{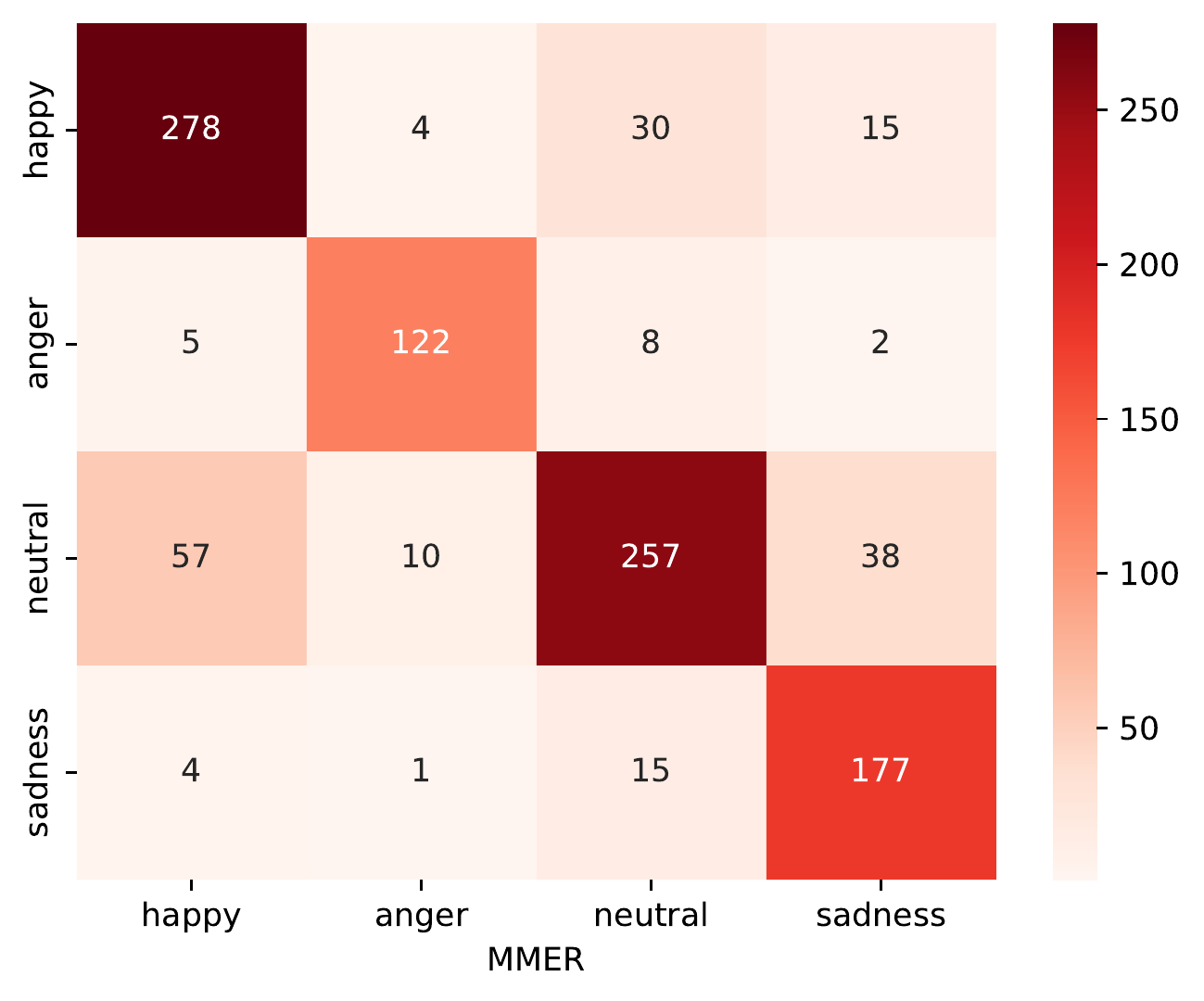}
\includegraphics[width=0.45\columnwidth]{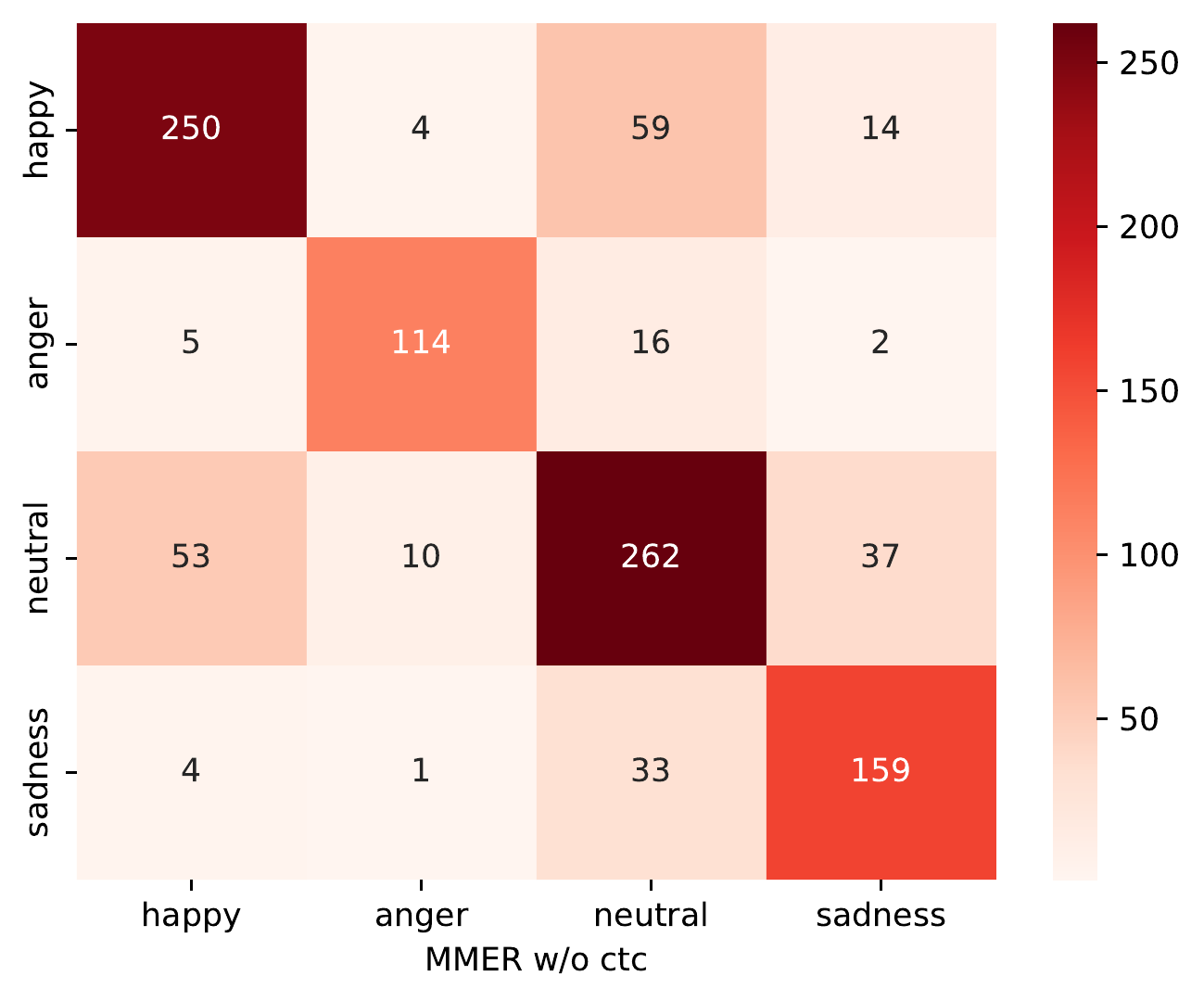}
\includegraphics[width=0.45\columnwidth]{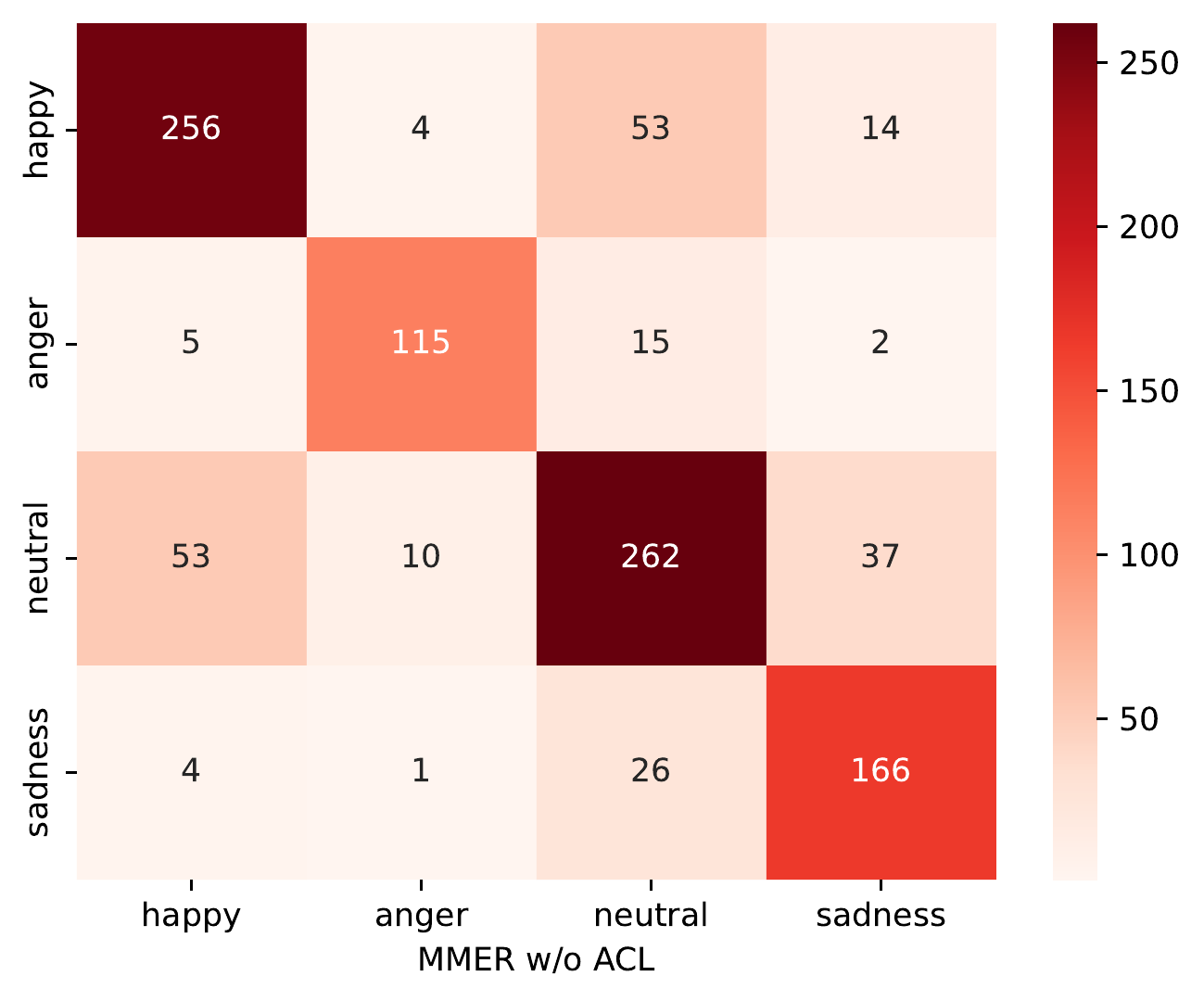}
\includegraphics[width=0.45\columnwidth]{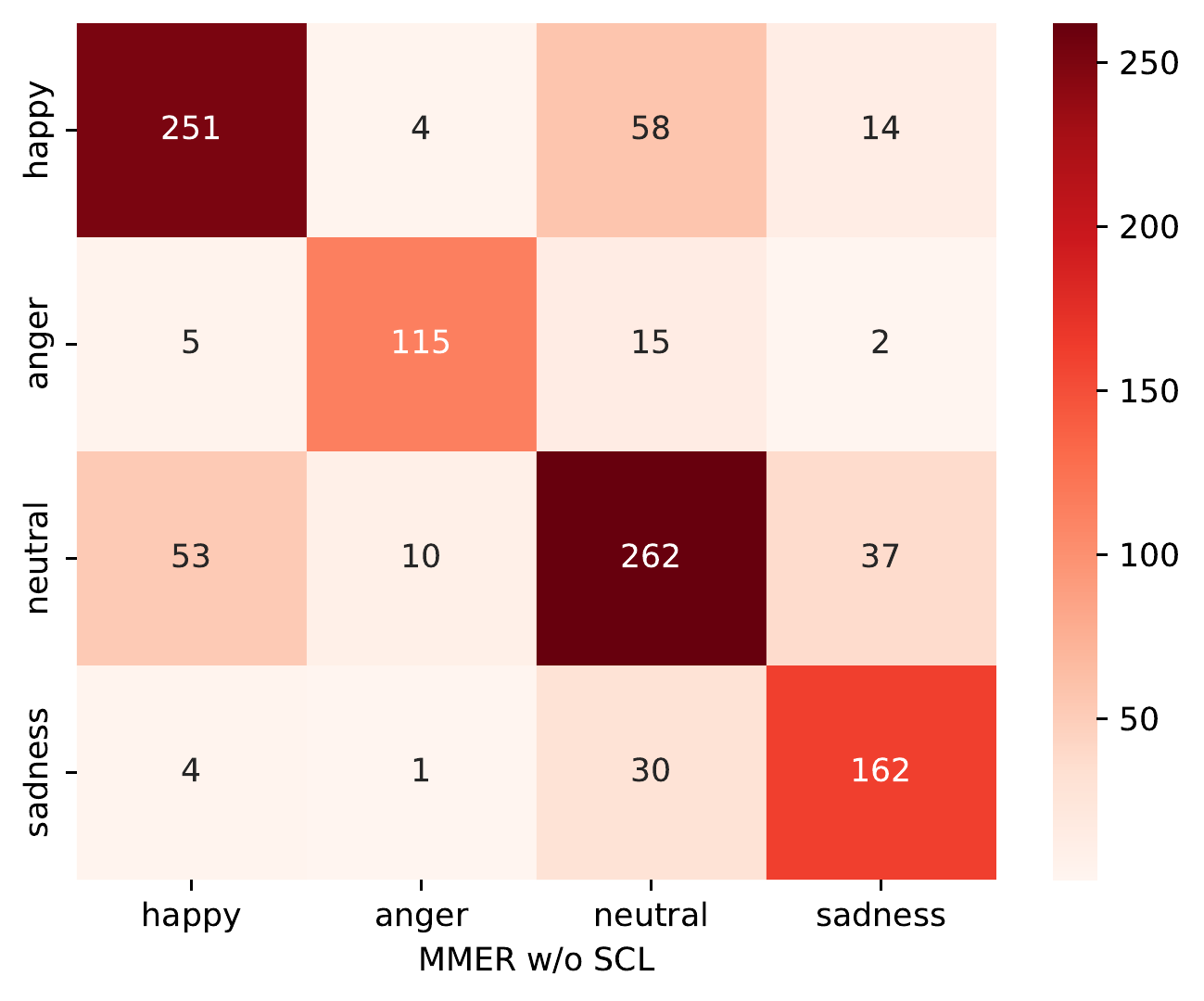}
\caption{\small Confusion Matrix: Ablation study and MMER performance evaluation with and without the CTC loss and the proposed objective functions, SCL and ACL..} 
\label{fig:cm83}
\end{figure}

{\noindent \textbf{Qualitative Analysis.}} Fig. \ref{fig:cm83} shows the confusion matrix for various settings with or without a particular objective function. One clear observation is that ACL alleviates the bias to the neutral class, which is a common problem in prior art \cite{keysparse}. On the contrary, CTC amplifies this by a small amount, which we attribute to the fact that our model learns more semantic information in text, ignoring important cues in speech. We provide results on various settings of $\alpha,\beta$, and $\gamma$ on our GitHub.



\setlength{\tabcolsep}{18pt}
{\renewcommand{\arraystretch}{1.125}%
\begin{table}[t]
\centering
\small
\caption{Emotion Recognition Results on IEMOCAP}
\scriptsize
\resizebox{0.95\columnwidth}{!}{
\begin{tabular}{l|ccc}
\toprule \toprule
\textbf{Method} &\textbf{CV} &\textbf{Modality} &\textbf{WA} \\
\midrule
\textbf{Prior-art} & & & \\
Wu et al. \cite{wuetal} &10-fold &\{a\} &72.7\% \\
Sajjad et al. \cite{sajjadetal} &5-fold &\{a\} &72.3\% \\
Lu et al. \cite{luetal1} &10-fold &\{a\} &72.6\% \\
Liu et al. \cite{liuetal2} &5-fold &\{a\} &70.8\% \\
Wang et al. \cite{wangetal} &5-fold &\{a\} &73.3\% \\
Zhao et al. \cite{zhao_22} &5-fold &\{a,t\} &76.3\% \\
Yang et al. \cite{yang22q_interspeech} &5-fold &\{a,t\} &77.7\% \\
Morais et al. \cite{morais_self} &5-fold &\{a,t\} &77.4\% \\
Chen et al. \cite{keysparse} &5-fold &\{a,t\} &74.3\% \\
Padi et al. \cite{padi2022multimodal} &5-fold &\{a,t\} &75.0\% \\
Makiuchi et al. \cite{9688036} &5-fold &\{a,t\} &73.5\% \\
Chen et al. \cite{chen2021key} &5-fold &\{a,t\} &74.3\% \\
Cai et al. \cite{cai2021speech} &10-fold &\{a,t\} &77.1\% \\
\midrule
\textbf{Our Baselines} & & & \\
RoBERTa\textsubscript{\emph{BASE}} &5-fold &\{t\} &69.2\%\\
wav2vec-2.0 &5-fold &\{a\} &73.9\% \\
Multimodal &5-fold &\{a,t\} &74.1\% \\
\midrule
\textbf{Proposed} & & & \\
MMER w/o CTC & 5-fold &\{a,t\} & 78.1\%\\
MMER w/o SCL & 5-fold&\{a,t\} & 78.9\%\\
MMER w/o ACL & 5-fold &\{a,t\} & 79.8\%\\
\textbf{MMER} & \textbf{5-fold} & \textbf{\{a,t\}} & \cellcolor{red!20}\textbf{81.2}\%\\
\bottomrule
\end{tabular}}
\label{tab:results}
\end{table}
}

\section{Conclusion and Limitations}
In this paper, we propose MMER, a novel multimodal multi-task approach for SER from spoken utterances. MMER leverages a novel dynamic multimodal fusion network and three additional auxiliary tasks. As part of future work, we would like to work on the current limitations of MMER, including the requirement of large batch sizes for training in contrastive learning and pre-computed text features.

\pagebreak
\bibliographystyle{IEEEtran}
\bibliography{mybib}

\end{document}